\documentclass[10pt,twocolumn,letterpaper]{article}

\usepackage{cvpr}              

\usepackage{multirow}
\usepackage{makecell}
\usepackage{amsmath}
\usepackage{amsfonts}

\usepackage{algorithm}
\usepackage{algorithmic}
\usepackage{booktabs}
\usepackage{graphicx}
\usepackage{algorithm}
\usepackage{algorithmic}

\usepackage{newfloat}
\usepackage{pifont}
\usepackage{listings}
\usepackage{xcolor}
\newfloat{listing}{tb}{lst}{}
\floatname{listing}{Listing}
\usepackage{threeparttable}
\usepackage{subcaption} 

\definecolor{cvprblue}{rgb}{0.21,0.49,0.74}
\usepackage[pagebackref,breaklinks,colorlinks,allcolors=cvprblue]{hyperref}

\title{CharGen: High Accurate Character-Level Visual Text Generation Model with MultiModal Encoder}

\author{Lichen Ma \;\;\; Tiezhu Yue \;\;\; Pei Fu \;\;\; Yujie Zhong \;\;\; Kai Zhou \;\;\; Xiaoming Wei \;\;\; Jie Hu\thanks{Corresponding author and project lead}\\
Meituan\\
{\tt\small\{malichen,yuetiezhu,fupei,zhongyujie,zhoukai03,weixiaoming,hujie39\}@meituan.com}
}

\begin{document}
\maketitle
\begin{abstract}
Recently, significant advancements have been made in diffusion-based visual text generation models. Although the effectiveness of these methods in visual text rendering is rapidly improving, they still encounter challenges such as inaccurate characters and strokes when rendering complex visual text.
In this paper, we propose CharGen, a highly accurate character-level visual text generation and editing model. Specifically, CharGen employs a character-level multimodal encoder that not only extracts character-level text embeddings but also encodes glyph images character by character. This enables it to capture fine-grained cross-modality features more effectively. Additionally, we introduce a new perceptual loss in CharGen to enhance character shape supervision and address the issue of inaccurate strokes in generated text.
It is worth mentioning that CharGen can be integrated into existing diffusion models to generate visual text with high accuracy. CharGen significantly improves text rendering accuracy, outperforming recent methods in public benchmarks such as AnyText-benchmark and MARIO-Eval, with improvements of more than $8\%$ and $6\%$, respectively. Notably, CharGen achieved a $5.5\%$ increase in accuracy on Chinese test sets.
\end{abstract}
\begin{table}[pbt] 
\centering                

\includegraphics[width=0.47\textwidth]{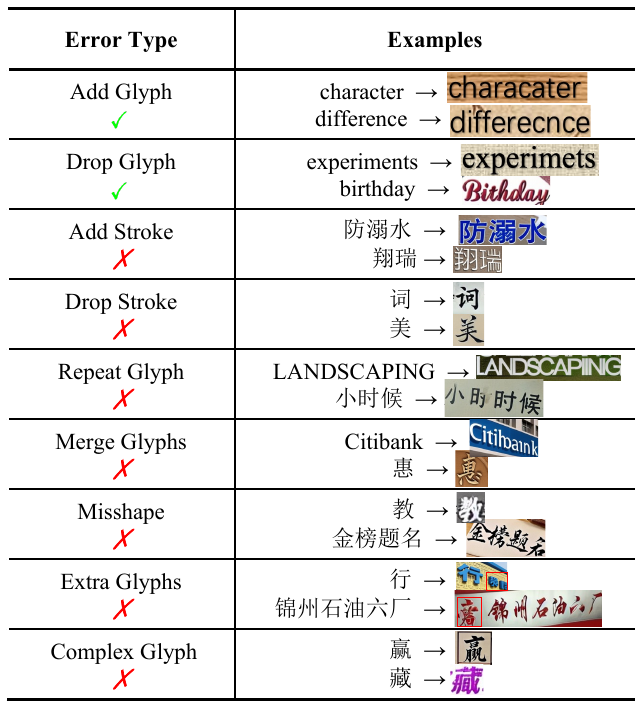}
\caption{Error types in visual text generation. (\textcolor{green}{\ding{51}}) means the error type solved by the previous methods, 
 (\textcolor{red}{\ding{55}}) means unsolved.}\label{error_type}  
\end{table}
\begin{figure*}[htbp]
    \centering
    \includegraphics[width=1.0\textwidth]{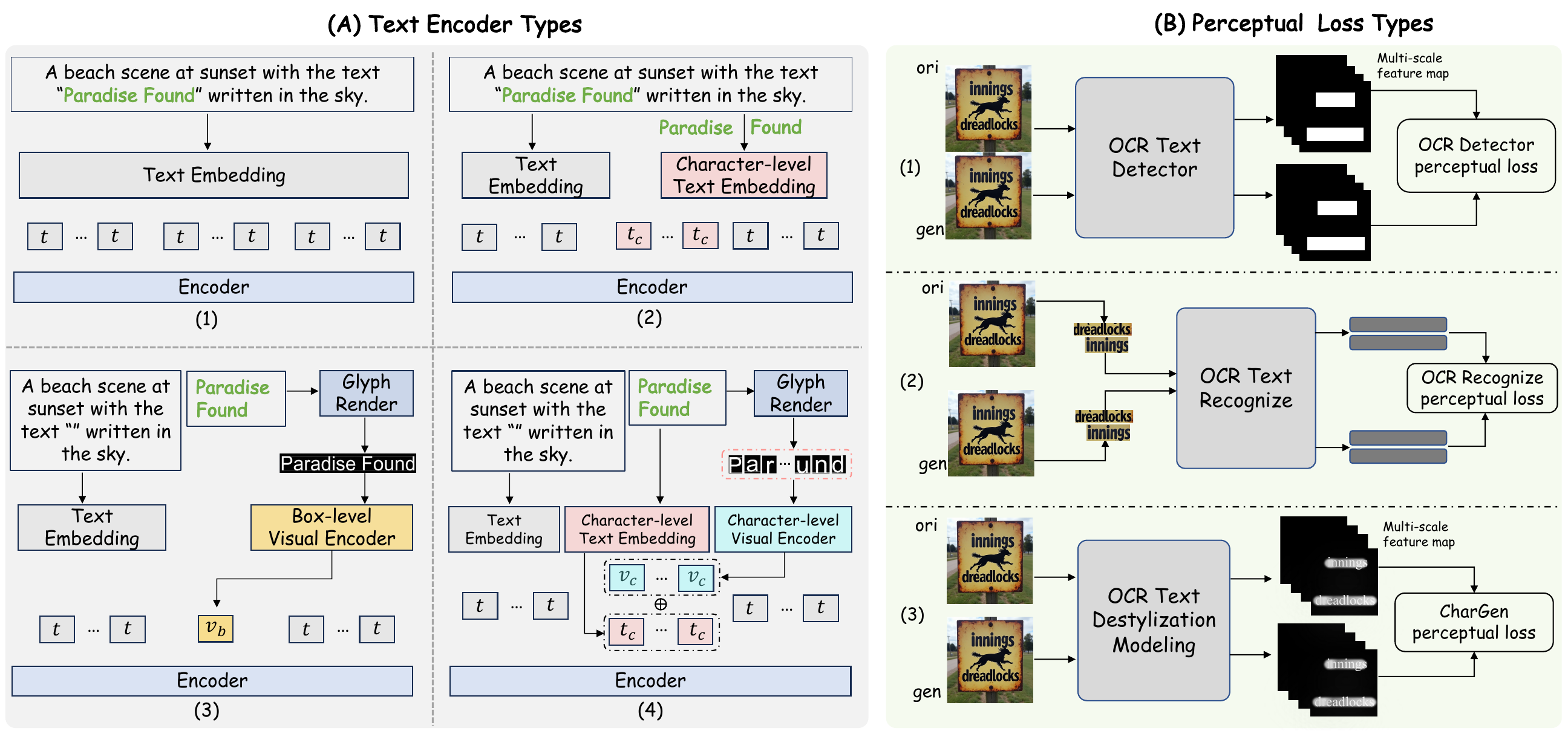}
    \caption{(A.1) A conventional text encoder without visual glyph information. (A.2) A character-level text embedding that indirectly incorporates glyph information. (A.3) A block-level visual embedding that replaces text embedding. (A.4) A character-level multi-modal encoder. (B.1) Based on an OCR detection model. (B.2) Based on an OCR recognition model. (B.3) Based on an ODM pre-trained model.}
    \label{encoder_loss}
\end{figure*}
\section{Introduction}
\label{sec:intro}

In recent years, text-to-image generation models based on diffusion models have seen rapid development. Notable advancements, such as Midjourney V6~\cite{midjourney}, DALL·E 3~\cite{betker_improving_nodate}, and SD3.0~\cite{esser_scaling_2024}, have demonstrated significant improvements in the field of text-to-image generation. Despite substantial progress in the image quality of generative models, these models still face challenges in generating accurate and readable visual text.
To address the need for refined and multi-styled visual text images, extensive research~\cite{chen_diffute_2023,zhao_udifftext_2023,zhang_brush_2023,ma_glyphdraw_2023,yang_glyphcontrol_2023,chen_textdiffuser_2023,chen_textdiffuser-2_2023,tuo_anytext_2024,liu2023characterawaremodelsimprovevisual} has been conducted based on open-source base models~\cite{ho_denoising_2020,ramesh2021zeroshottexttoimagegeneration,song2021scorebasedgenerativemodelingstochastic,dhariwal_diffusion_2021,nichol2021improveddenoisingdiffusionprobabilistic,saharia_photorealistic_2022,ramesh2022hierarchicaltextconditionalimagegeneration,rombach_high-resolution_2022,chang2023musetexttoimagegenerationmasked}, primarily focusing on visual text generation. Compared to open-source general models, these methods can accomplish simple visual text generation tasks. However, previous methods still need to improve the accuracy of generated visual text, which is a crucial factor in various image generation applications.

The visual text generation framework can be divided into three key components: \textit{text encoder}, \textit{control conditions}, and \textit{perceptual supervision}. Previous methods have primarily focused on the design of the text encoder and perceptual supervision components, demonstrating that improvements in these areas can significantly enhance the accuracy of text generation. However, these methods still encounter persistent errors in English and Chinese text rendering, such as Repeat Glyph, Merge Glyphs, and Misshape errors, as illustrated in Table~\ref{error_type}.

In response to these challenges, we propose a new visual text generation framework, CharGen. This framework aims to produce images with well-structured, clear, and coherent text by advancing the text encoder and perceptual supervision components. CharGen addresses the limitations of previous methods and strives to deliver higher accuracy and quality in visual text rendering.

For the text encoder component, we introduce a character-level multimodal encoder that incorporates fine-grained multimodal features into the control conditions, as illustrated in Fig.~\ref{encoder_loss}~(A.4). This encoder operates in two main ways: firstly, it employs a visual encoder to process rendered glyph images character by character, thereby guiding text generation at the character level. Secondly, it integrates character-level text embeddings to retain semantic information as much as possible while effectively distinguishing visually similar characters.

Regarding the perceptual supervision component, inspired by the ODM (OCR-Text Destylization Modeling) approach~\cite{duan_odm_2024}, we propose an auxiliary perceptual loss specifically designed for visual text generation, termed CharGen Loss, as depicted in Fig.~\ref{encoder_loss}~(B.3). This loss function supervises the visual text glyphs from both the generated image and the original image, considering the entire image perspective as well as a more fine-grained approach. By leveraging CharGen Loss based on the destylization encoder, the training process can minimize the constraints imposed by font styles. This ensures that the generated text style is more influenced by the descriptive text of the image, thereby maintaining consistency between the text style and the overall scene style.

In summary, our contributions are as follows:
\begin{itemize}
    \item We introduce CharGen, a new high-precision visual text generation framework based on ControlNet~\cite{zhang_adding_2023_arx}, to address the errors in visual text generation tasks, as outlined in Table~\ref{error_type}.
    \item Within CharGen, we propose a character-level multimodal encoder that incorporates fine-grained multimodal features into the control conditions. Additionally, we introduce a multi-scale perceptual loss based on the ODM pre-trained model, which provides more accurate supervision of visual text glyphs.
    \item Our method demonstrates significant improvements over previous methods in terms of OCR accuracy, NED, and FID, achieving state-of-the-art results on both Chinese and English benchmarks.
\end{itemize}

\begin{figure*}[htbp]
    \centering
    \includegraphics[width=0.94\textwidth]{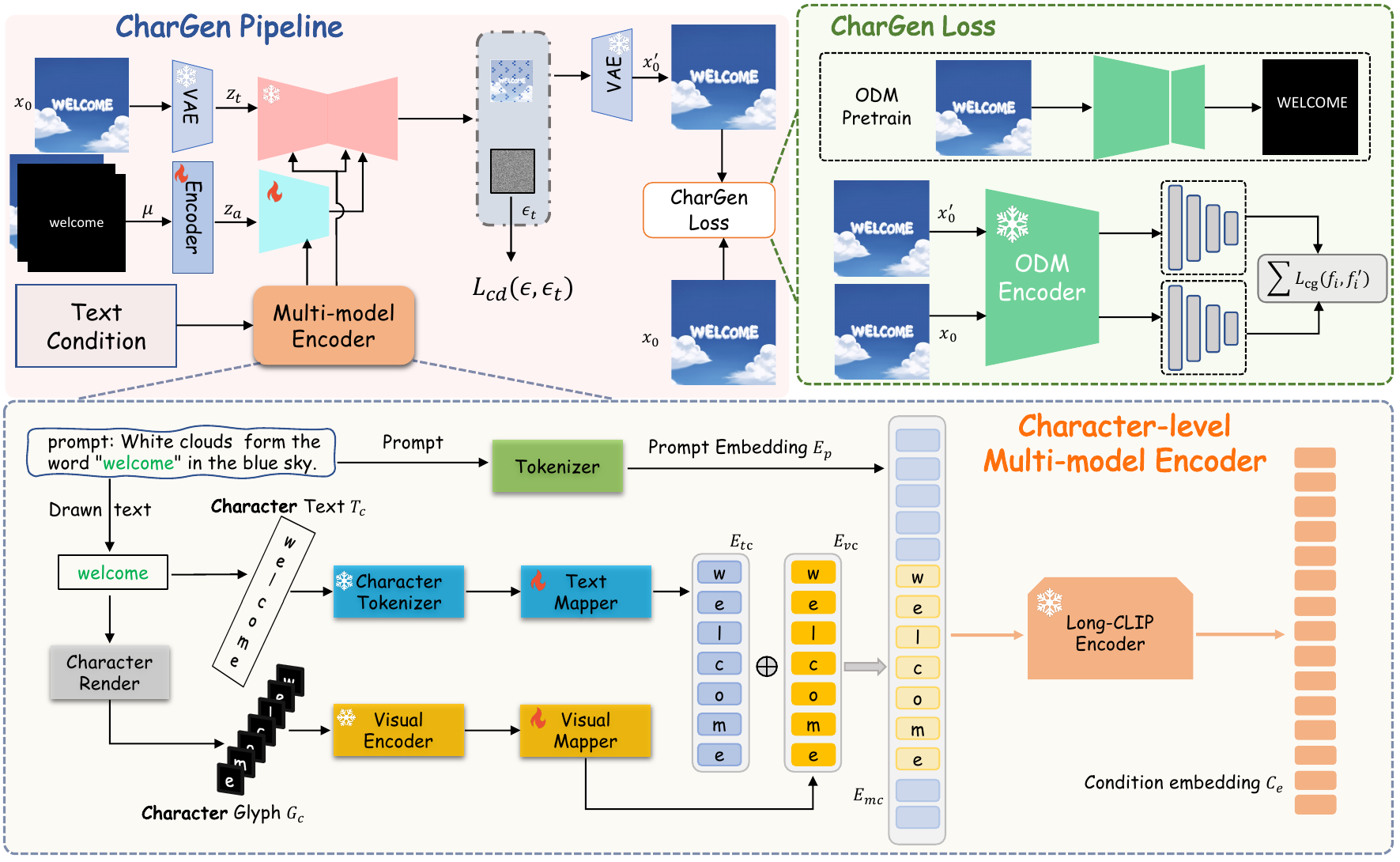}
    \caption{The framework of CharGen.}\label{main}
\end{figure*} 

\section{Related work}
\label{sec:formatting}
\textbf{Visual Text Generation.} Over the last few years, significant advancements have been made in the field of visual text generation, as demonstrated by various methods~\cite{chen_diffute_2023,zhao_udifftext_2023,zhang_brush_2023,ma_glyphdraw_2023,yang_glyphcontrol_2023,chen_textdiffuser_2023,chen_textdiffuser-2_2023,tuo_anytext_2024}. These models have evolved beyond merely generating simple visual text images, leading to developments in interactive image editing. Recent research has focused on three critical aspects of visual text generation: control condition, text encoder, and perceptual supervision.

\textbf{Control Condition.} To guide text generation or editing, auxiliary glyph images of the text to be generated or edited are commonly used. For instance, TextDiffuser~\cite{chen_textdiffuser_2023} initially generates segmentation masks of characters and then utilizes these character segmentation masks along with position masks to facilitate image generation. Similarly, GlyphControl~\cite{yang_glyphcontrol_2023} employs ControlNet~\cite{zhang_adding_2023_arx} to encode glyph images. In our work, we adopt a similar strategy to GlyphControl for rendering glyph images, while also incorporating position and mask images as additional conditions.

\textbf{Text Encoder.} We summarize the implementation of previous methods in the text encoder component in Fig.~\ref{encoder_loss}~(A.1-A.3). Most base models use a text encoder without incorporating any visual glyph information. DrawText~\cite{liu_character-aware_2023} argues that character-blind text encoders (e.g. CLIP~\cite{radford2021learningtransferablevisualmodels}, and T5~\cite{raffel2023exploringlimitstransferlearning}) typically used for image generation lack core spelling knowledge, making it challenging to translate input tokens into glyph sequences.
To address this, UDiffText~\cite{zhao_udifftext_2023} and Glyph-ByT5 (v1, v2)~\cite{liu_glyph-byt5_2024, liu_glyph-byt5-v2_2024} implicitly inject glyph information by retraining the text encoder through contrastive learning. In DiffUTE~\cite{chen_diffute_2023} and GlyphDraw~\cite{ma_glyphdraw_2023}, glyph images are directly integrated into the text embedding. DiffUTE extracts glyph information from an OCR recognition model, while GlyphDraw takes the off-the-shelf CLIP image encoder to extract glyph embeddings. However, CLIP is not a good choice for extracting glyph information. It is only suitable for extracting high-level semantic features, whereas glyphs require low-level image structural features.
AnyText~\cite{tuo_anytext_2024} and GlyphDraw2~\cite{ma2024glyphdraw2} render a glyph line with multiple characters into an image (as shown in Fig.~\ref{encoder_loss}~(A.3)), encode glyph information using a pretrained OCR recognition model, and inject it into the text embedding by replacing its embedding from a special token embedding in the caption. Unfortunately, the glyph line with multiple characters in a token is difficult to deconstruct into specific glyph information. 
We propose a method where, during the text embedding process, glyph embedding is performed character by character while retaining the original character's text embedding encoding. Subsequently, the image modality encoding and text modality encoding are fused together to serve as the characters' representation.

\textbf{Perceptual Supervision.} Perceptual supervision has been widely employed for glyph supervision. We summarize several perceptual losses in Fig.~\ref{encoder_loss}. OCR-VQGAN~\cite{rodriguez2023ocr} leverages a pre-trained OCR detection model to extract image features and supervises text generation by constraining discrepancies across multiple intermediate layers. TextDiffuser~\cite{chen_textdiffuser_2023} employs a character-level segmentation model to ensure the precision of each generated character in the latent space. AnyText~\cite{tuo_anytext_2024} adopts an OCR recognition model proficient in stroke and spelling to supervise text generation exclusively within the specified text region.
As shown in Fig~\ref{encoder_loss}~(B.2), the encoder based on OCR recognition loses the spatial position information of the entire image. Although the encoder based on OCR detection in Fig.~\ref{encoder_loss}~(B.1) retains the spatial position information, it lacks fine-grained supervision of glyphs.

To address these limitations, CharGen utilizes the image encoder of a pre-trained ODM (OCR-Text Destylization Modeling)~\cite{duan_odm_2024} to supervise the destylized glyphs, thereby avoiding style constraints. This approach provides a more direct and effective form of supervision, ensuring accurate and diverse visual text generation.

\section{Method}
In this section, we will delineate the framework of CharGen through three distinct parts. Firstly, the preliminary part will provide a concise overview of the backbone network, ControlNet, along with the auxiliary latent inputs. Secondly, the character-level multimodal encoder part will elaborate on the methodology for effectively integrating character-level text and visual modalities within the text encoder. Lastly, we will detail the multi-scale perceptual loss, which is based on the pre-trained ODM model.

\subsection{Preliminary}

Our work builds upon the ControlNet architecture~\cite{zhang_adding_2023}, utilizing SD1.5~\cite{rombach_high-resolution_2022} for model weight initialization. ControlNet is a robust network that enhances pre-trained large diffusion models by incorporating additional input conditions. It is capable of learning task-specific conditions in an end-to-end manner, even with limited training data. In this work, we categorize conditions into two types: image-based conditions and text conditions.

As depicted in Fig.~\ref{main}, both the original image feature and the mask image feature are embedded using VAE~\cite{kingma2022autoencodingvariationalbayes}. To derive the image-based conditions, we apply several convolutional layers to encode glyph and position information, subsequently fusing these features into $z_a$. This $z_a$ is combined with a noisy latent image $z_t$ and then fed into ControlNet as image-based conditions.

For text conditions, we have innovatively developed a multimodal character-level encoding module to extract visual and text features, which serve as text conditions. This module will be discussed in detail in Subsection~\ref{Multimodal encoder}. Additionally, we incorporate a masked image, inspired by AnyText, to manage the image area that should be preserved during the diffusion process. This approach also allows us to control the transition between image generation and image editing based on the probability $\mu$.

\begin{figure*}[t]
    \centering
    \includegraphics[width=0.9\textwidth]{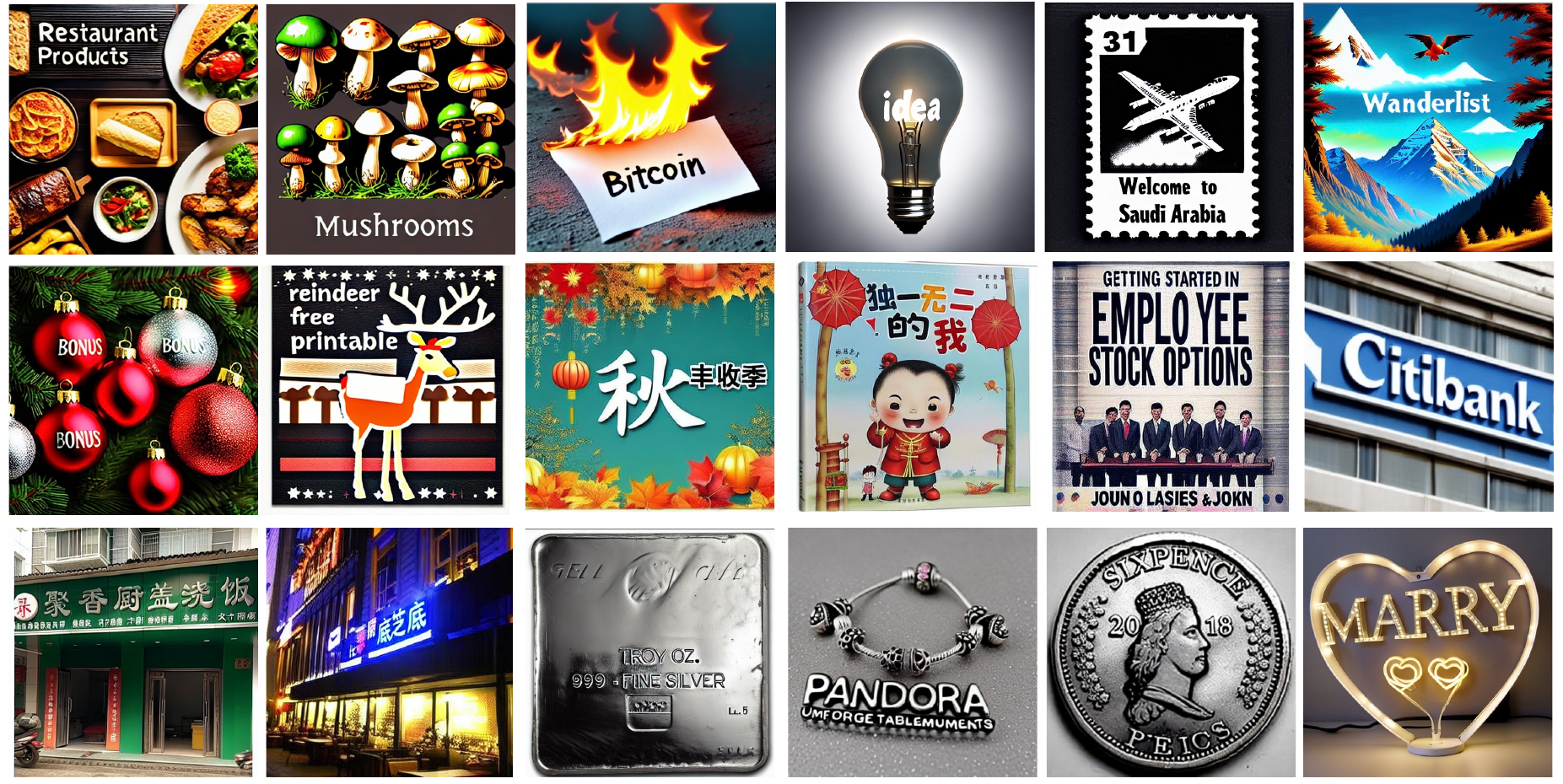}
    \caption{Visual text image generated by CharGen.}\label{en-compare-img}
\end{figure*} 

\subsection{Character-level Multimodal Encoder}\label{Multimodal encoder}
Generating accurate multilingual text using diffusion models presents significant challenges, particularly when dealing with complex words and multi-stroke characters. Our research indicates that existing methods often encounter issues such as inaccuracies in character formation and stroke rendering when generating complex visual text. This is largely due to the text encoder's limitations in incorporating rich, fine-grained information, which hampers its ability to effectively guide text generation. To overcome these challenges, we propose a character-level multimodal encoder designed to extract more precise text conditions.
 

We will now introduce the multimodal embedding module in detail. Initially, a single character is rendered into a character glyph  $G_c$. For instance, a word or sentence is rendered using a uniform font style (Arial Unicode) as a group of glyphs $G_c = \{G_c^1, G_c^2, \cdots, G_c^{n-1}, G_c^n\}$. Subsequently, a character-by-character visual encoder extracts the visual embeddings $E_{vc} = \{E_{vc}^1, E_{vc}^2, \cdots, E_{vc}^{n-1}, E_{vc}^{n}\}$ from the glyph images $G_c$. Concurrently, a text encoder is employed to extract character-level text embeddings $E_{tc} = \{E_{tc}^1, E_{tc}^2, \cdots, E_{tc}^{n-1}, E_{tc}^{n}\}$. 
We propose a character-level visual feature mapper module and a text feature mapper module based on two decoupled linear layers. These mappers are used to fuse the text embeddings $E_{tc}$ and visual embedding $E_{vc}$ into multimodal embeddings $E_{mc}= \{E_{mc}^1, E_{mc}^2, \cdots, E_{mc}^{n-1}, E_{mc}^{n}\}$. Simultaneously, the prompt is encoded into token-sized embeddings $E_p = \{E_p^1, E_p^2, \cdots, E_p^{m-1}, E_p^{m}\}$. The prompt embeddings and multimodal embeddings are combined into a text condition $E_{pm} = \{E_p, E_{mc}\}$, which is then fed into a transformer-based text encoder to obtain a fused intermediate representation. This representation is subsequently mapped to the intermediate layer of UNet using a cross-attention mechanism.
The inclusion of the text embedding $E_{tc}$ ensures that the condition can carry the semantic features of the text, thereby maintaining the integrity of the multimodal embedding $E_{mc}$. Additionally, compared to using only the visual module, we found that the multimodal character-level embedding extraction module better addresses issues of inaccurate characters and strokes when generating complex words and multi-stroke characters.
 
The upper limit for the number of text tokens in the CLIP text encoder is 77. Previous studies have demonstrated that its actual effective length is often less than 20 tokens. This limitation hampers CLIP's ability to handle detailed descriptions, thereby restricting the model's understanding and application capabilities. To address this issue and increase the capacity for text generation, we adopted the latest Long-CLIP~\cite{zhang2024longclip} scheme, which raises the number of text tokens from 77 to 248. Long-CLIP not only supports longer text input but also maintains or even exceeds CLIP's zero-shot generalization ability. Furthermore, it aligns with the CLIP latent space, enabling seamless integration into downstream frameworks without requiring additional adaptation.

\begin{figure*}[t]
    \centering
     \includegraphics[width=0.9\textwidth]{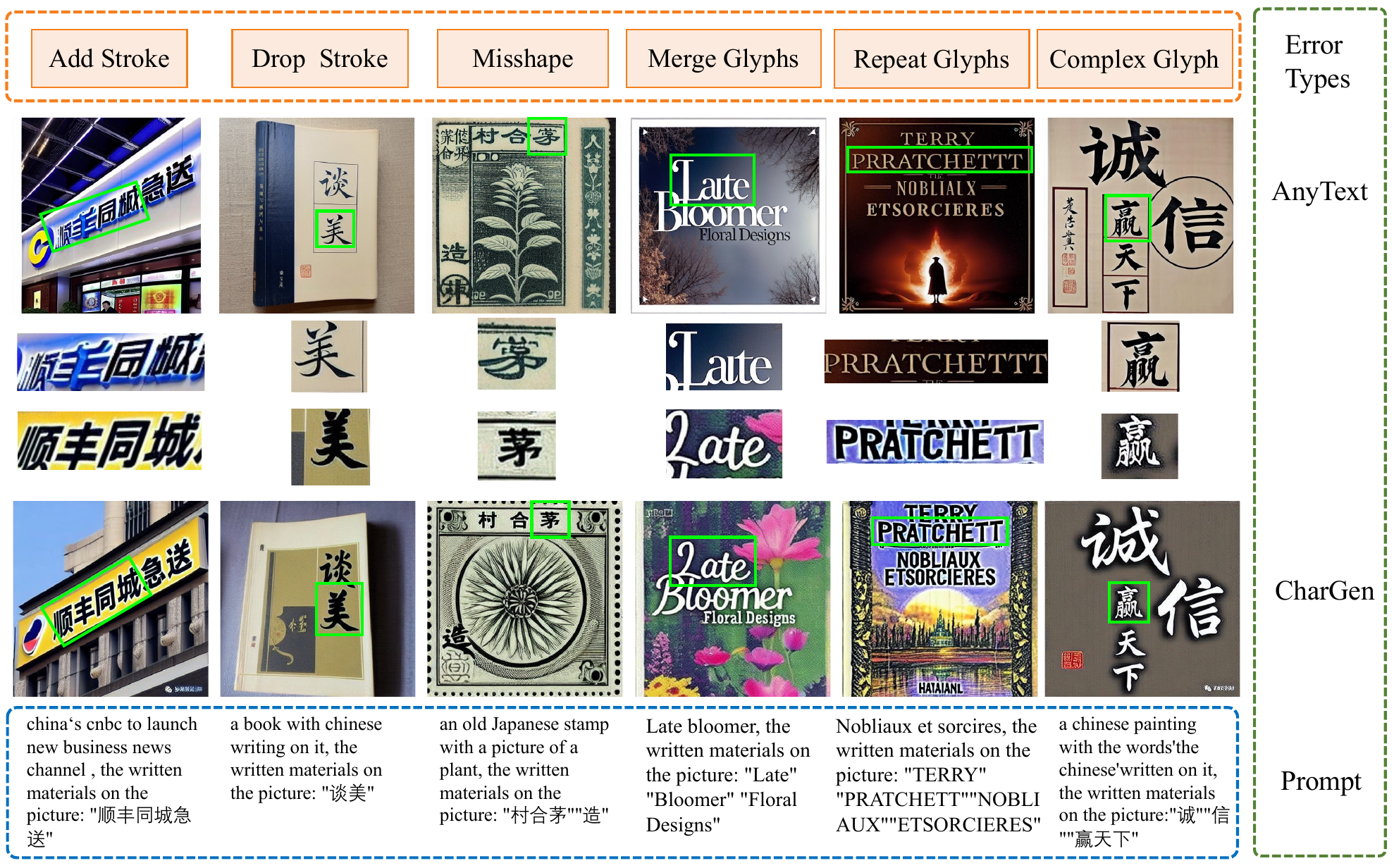}
    \caption{A qualitative comparison of CharGen with AnyText on English and Chinese text generation, using test captions from the AnyText-benchmark dataset.}\label{any-bench}
\end{figure*}

\subsection{Perceptual Loss}
Previous methods often faced challenges in generating accurate strokes for complex words, resulting in outputs that were either unreadable or of low quality to users. We believe this issue arises from the use of inaccurate glyph loss supervision, where the level of supervision granularity is insufficiently fine to effectively monitor tiny fonts. For instance, backbone networks like OCR emphasize robustness during training and can recognize the same word even with missing or added strokes, thus overlooking errors in glyphs.

To address this problem, we proposed a multi-scale perceptual loss based on the ODM to enhance fine-grained text glyph generation. Unlike previous training objectives, such as text detection and text bounding box area segmentation, the pre-training objective of the ODM network is specifically tailored to restore text glyphs, ignoring features from other areas. This approach aligns with our goal for glyph supervision, which is to concentrate solely on text glyphs while disregarding color, texture, and other extraneous object information.

During the training phase, the denoiser network predicts noise, and we use the noisy latent image to predict the features of the latent space. We further employ the VAE decoder to approximate the image and convert it from the latent space to the image space. We input predicted image $I_{pred}$ and original image $I_{ori}$ into the ODM encoder model to extract multi-scale features, denoted as $\hat{f}_{hw}, f_{hw} \in \mathbb{R}^{{H_k}\times{W_k}\times{C_k}}$ for each layer $k$. The CharGen perceptual loss is formulated as follows:\\
\begin{equation}
f_{hw} = ODM(I_{ori}), \hat{f}_{hw} = ODM(I_{pred}),
\end{equation}
\begin{equation}
\mathcal{L}_{chargen} = \sum^{k} \frac{1}{H_{k}W_k} \sum^{h,w} \parallel f_{hw}^k - \hat{f}_{hw}^k \parallel^{2}_2\label{XX}.
\end{equation}
Instead of employing a block-by-block calculation for OCR perceptual loss, we average over the spatial dimensions and sum across the channel dimension. Concurrently, the condition control diffusion loss is given by:
\begin{equation}
\mathcal{L}_{cd} = \parallel \epsilon_\theta(z_t,z_a,C_e,t) - \epsilon \parallel^{2}_2 \label{XX}.
\end{equation}
The overall training loss is defined as:
\begin{equation}
\mathcal{L} = \mathcal{L}_{cd} \label{XX} + \lambda * \mathcal{L}_{chargen},
\end{equation}
where $\lambda$ is a hyperparameter used to adjust the weight ratio between the two loss functions.\\

\section{Experiment}
\subsection{Implementation Details}
Our model is trained on the AnyWord-3M dataset for 20 epochs using 16 Tesla A100 GPUs. During training phase, we use AdamW optimizer with a learning rate of 2e-5 and a global batch size of 32. Further, we employ a progressive training strategy, initially disabling the edit branch for the first 10 epochs. In the subsequent 5 epochs, the edit branch is activated with a probability of  $\mu = 0.5$. During the final 5 epochs, the CharGen loss is incorporated into the training process with a weight coefficient of $\lambda = 0.01$. Each character is consistently rendered into an $80 \times 80$ glyph image $C_g$ with a font size of 60. Both the long-clip text embedding and visual features are mapped to a dimension of 768. 


CharGen is based on ControlNet, using SD1.5 as the initialization parameter. The text encoder employs long-clip, with the upper limit for encoding tokens set to 248. For feature extraction during training, the character-wise glyph visual encoder uses PP-OCRv3~\cite{li2022ppocrv3}. To ensure fairness in evaluation, we use another open-source model, DuGuangOCR~\cite{ModelScope}, as a reference, in line with AnyText.

\begin{table*}[h]
    \centering
    \begin{threeparttable}
    \begin{tabular*}{0.69\linewidth}{l|c|c|c|c|c|c}
    \toprule [1.5pt]
    \multirow{2}*{Methods}  & \multicolumn{3}{c|}{English} & \multicolumn{3}{c}{Chinese} \\
    \cline{2-4} \cline{5-7}
    & Sen.ACC $\uparrow$ & NED $\uparrow$ & FID $\downarrow$ & Sen. ACC $\uparrow$ & NED $\uparrow$ & FID $\downarrow$\\
    \midrule [1.5pt]
    ControlNet & 0.5910 & 0.8061 & 49.43 & 0.3727 & 0.6355 & 52.64  \\  
    TextDiffuser & 0.5878 & 0.7958 & 43.91 & 0.0600 & 0.1248 & 52.55  \\
    GlyphControl\dag & 0.3382 & 0.6390 & 36.49 & 0.0301 & 0.0775 & 32.04  \\
    GlyphControl\ddag & 0.4933 & 0.7324 & 43.35 & 0.0412 & 0.0939 & 48.17  \\
    AnyText\textbf{-v1.1} & 0.7222 & 0.8789 & \textbf{33.92} & 0.6943 & 0.8410 & 31.63  \\
    
    CharGen & \textbf{0.8096} & \textbf{0.9205} & 34.09 & \textbf{0.7499} & \textbf{0.8609} & \textbf{29.49}  \\
    \bottomrule [1.5pt]
    \end{tabular*}
    \end{threeparttable}
    \caption{Comparison on AnyText-benchmark dataset. $\dag$ indicates the model is trained on LAION-Glyph-10M, while $\ddag$ signifies it is fine-tuned on TextCaps-5k.} \label{tab:anytext}
\end{table*}


\begin{table} [pbt]
    \centering
    \begin{threeparttable}
    \begin{tabular*}{1.03\linewidth}{l|c|c|c|c}
    \toprule [1.5pt]
    \multirow{2}*{Methods} & Img. & Sen. & Sen. & Sen.  \\
        ~ & ACC & Precision & Recall & F-measure \\
    \midrule [1.5pt]
    TextDiffuser & 0.561  & 0.785  & 0.780  & 0.782  \\
    TextDiffuser2 & 0.576  & 0.740  & 0.761  & 0.751   \\
    Glyph-ByT5  & 0.748  & 0.882  & 0.926  & 0.904   \\
    CharGen  &\textbf{0.808 } & \textbf{0.928 } & \textbf{0.935 } & \textbf{0.931 } \\
    \bottomrule [1.5pt]
    \end{tabular*}
    \end{threeparttable}
    \caption{Comparison on MARIO-Eval dataset.}\label{tab:mario}
\end{table}

\begin{figure*}[htbp]
    \centering
    \includegraphics[width=0.95\textwidth]{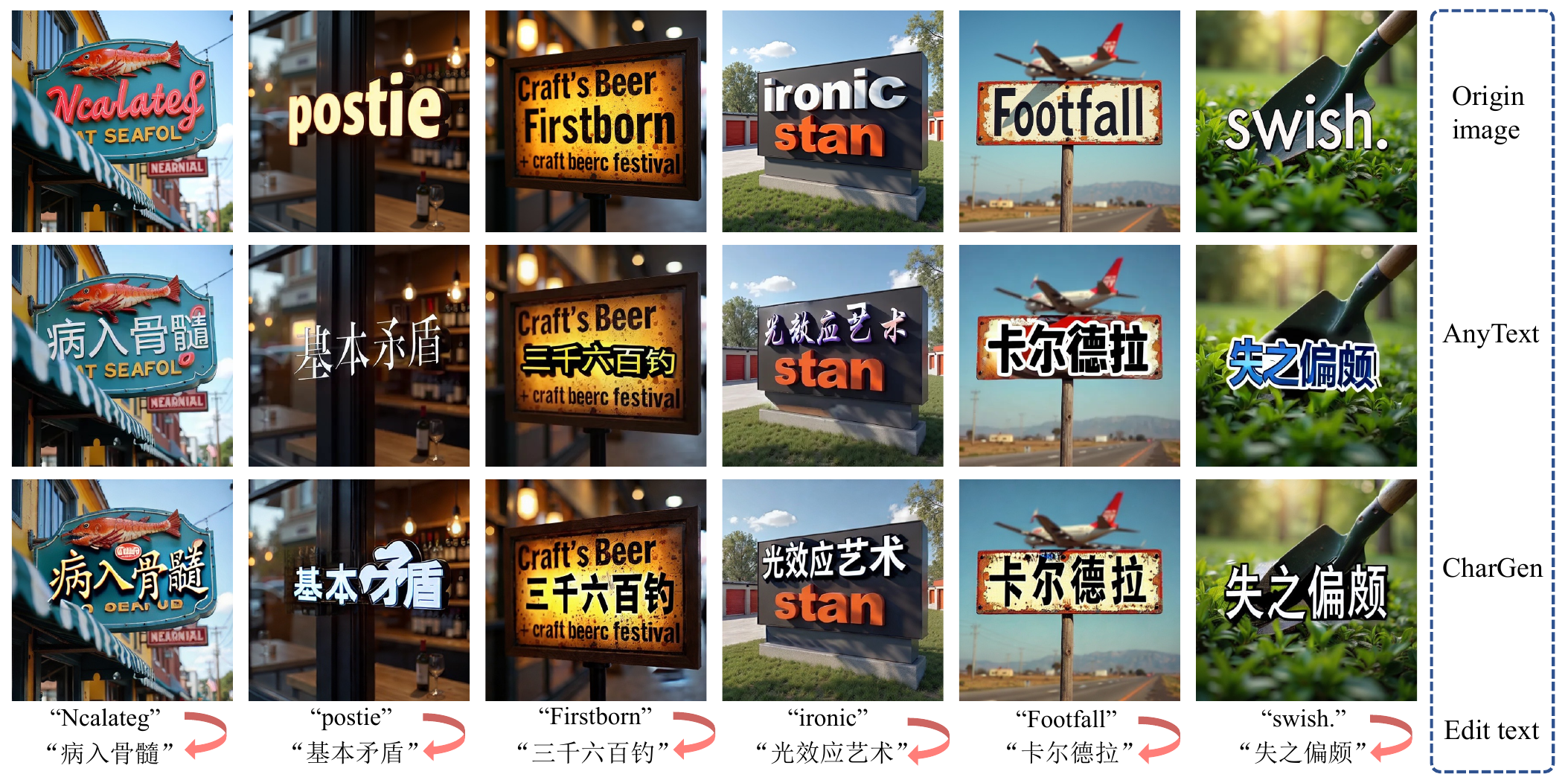}
    \caption{A qualitative comparison of CharGen with AnyText on visual text editing.}\label{edit-img}
\end{figure*} 

\subsection{Training Dataset and Benchmark}
\subsubsection{Training Dataset}
We use AnyWord-3M as the training dataset, which is a large-scale multilingual dataset of public images proposed by AnyText~\cite{tuo_anytext_2024}. The sources of these images include Noah-Wukong~\cite{gu2022wukong}, LAION-400M~\cite{schuhmann2021laion}, and various datasets designed for OCR recognition tasks. These images encompass a wide array of scenes containing text, such as street scenes, book covers, advertisements, posters, and movie frames, totaling 3.03 million images.

\subsubsection{Benchmark}
To evaluate CharGen and previous state-of-the-art methods, we employ the AnyText-benchmark and MARIO-Eval. The AnyText-benchmark includes test data in both Chinese and English, using the Sentence Accuracy (Sen.ACC) metric. For each generated visual text, the text is cropped according to the specified position and input into the OCR model to obtain the prediction result. The predicted text is deemed correct only if it exactly matches the ground truth. Additionally, we use the Normalized Edit Distance (NED) metric to calculate the similarity between the predicted string and the ground truth.
MARIO-Eval is a comprehensive benchmark collected from six different sources. Within the MARIO-Eval benchmark, we use image-level accuracy (Img.ACC) to assess the text generation model, and also evaluate the sentence-level precision, recall, and F-measure results.

\begin{table*} [h]
    \label{tab_dwsc}
    \centering
    \begin{tabular*}{0.82\linewidth}{c|c|c|c|c|c|c|c|c}
    \toprule [1.5pt]
    \multirow{2}*{Exp.} & \multirow{2}*{\makecell[c]{Embedding\\Granularity}} & \multicolumn{2}{c|}{Embedding Modal} & \multicolumn{2}{c|}{Perceptual Loss} & \multirow{2}*{$\lambda$} & \multirow{2}*{Sen. ACC $\uparrow$} & \multirow{2}*{NED $\uparrow$} \\
    \cline{3-4} \cline{5-6} 
    & & Visual & Text & OCR & ODM & &  \\
    \midrule [1.5pt]
    1 & Box-Level & \ding{51} & \ding{56} &  \ding{56} & \ding{56} & $ - $ & 0.460 & 0.707 \\ 
    2 & Box-Level & \ding{51} & \ding{56} & \ding{51} & \ding{56} & $ 0.01 $ & 0.500 & 0.746 \\ 
    3 & Box-Level & \ding{51} & \ding{56} & \ding{56} & \ding{51} & $ 0.01 $ & 0.524 & 0.759 \\ 
    4 & Character-Level & \ding{56} & \ding{51} &  \ding{56} & \ding{56} & $ - $ & 0.165 & 0.421 \\ 
    5 & Character-Level & \ding{51} & \ding{56} & \ding{56} & \ding{56} & $ - $ &  0.535 & 0.773 \\ 
    6 & Character-Level & \ding{51} & \ding{56} & \ding{51} & \ding{56} & $ 0.01 $ & 0.586 & 0.798 \\ 
    7 & Character-Level & \ding{51} & \ding{56} & \ding{56} & \ding{51} & $ 0.01 $ & 0.620 & 0.809\\ 
    8 & Character-Level & \ding{51} & \ding{51} & \ding{56} & \ding{56} & $ - $ & 0.572 & 0.791\\ 
    9 & Character-Level & \ding{51} & \ding{51} & \ding{56} & \ding{51} & $ 0.01 $ & \textbf{0.639} & \textbf{0.813}\\ 
    10 & Character-Level & \ding{51} & \ding{51} & \ding{56} & \ding{51} & $ 0.005 $ & 0.635 & 0.807\\ 
    11 & Character-Level & \ding{51} & \ding{51} & \ding{56} & \ding{51} & $ 0.03 $ & 0.627 & 0.801\\ 
    \bottomrule [1.5pt]
    \end{tabular*}
    \caption{Ablation experiments of CharGen conducted on a subset of the AnyWord-3M dataset.} \label{tab:ablation strudy}
\end{table*}

\subsection{Results}
\subsubsection{Quantitative Comparison Results} 

We utilize the AnyText benchmark and MARIO-Eval to assess previous methods, including ControlNet~\cite{zhang_adding_2023}, TextDiffuser~\cite{chen_textdiffuser_2023}, TextDiffuser-2~\cite{chen_textdiffuser-2_2023}, GlyphControl~\cite{yang_glyphcontrol_2023}, GlyphDraw2~\cite{ma2024glyphdraw2}, AnyText~\cite{tuo_anytext_2024}, and Glyph-ByT5~\cite{liu_glyph-byt5_2024}. For a fair evaluation, in line with AnyText, all methods employ the DDIM sampler with 20 sampling steps, a CFG-scale of 9, a fixed random seed of 100, and identical positive and negative prompt words.

Quantitative comparison results with previous methods on the AnyText-benchmark are presented in Table~\ref{tab:anytext}. From these results, CharGen achieves state-of-the-art performance on both Chinese and English public benchmarks compared to previous methods. Although we selected AnyWord-3M as the training data, which includes only 1.39 million English samples, TextDiffuser and GlyphControl were trained on 10 million pure English data.

In comparison with previous methods on the AnyText-benchmark dataset, as illustrated in Table~\ref{tab:anytext}, our method shows an improvement of $8.8\%$ on the English dataset, achieving $80.96\%$ on the Sen.ACC metric compared to AnyText. For Chinese text generation, CharGen improved by $5.5\%$, reaching $74.99\%$ on the Sen.ACC evaluation metric compared to AnyText. By examining the text-generated images, we found that our method excels in generating text glyphs, particularly for multi-stroke characters and similar characters. This advantage is attributed to the character-level multimodal encoder we proposed, which better addresses errors such as repeated glyphs and added/dropped strokes in visual text generation compared to previous methods.
Moreover, by employing character-level multimodal embeddings and CharGen loss, our method improves the accuracy of generated text shapes and resolves challenging issues such as misshaped and merged glyphs, as detailed in Table~\ref{error_type}.

Compared with the previous state-of-the-art method, Glyph-ByT5, on the MARIO-Eval dataset, CharGen demonstrates significant improvements over previous methods at both image and sentence levels, as shown in Table~\ref{tab:mario}. An interesting observation is that both CharGen and Glyph-ByT5 exhibit much better performance in terms of image-level accuracy when trained with synthetic-text data compared to scene-text data. This suggests that the controlled environment and clarity of synthetic-text data may contribute to more effective training and better model performance.

\subsubsection{Qualitative Comparison Results}
Next, we analyze various error types of previous methods in Table~\ref{error_type} and conduct a qualitative comparison with AnyText. For English visual text generation, AnyText encountered difficulties with merged glyphs and repeated glyphs when generating complex long texts. In contrast, CharGen effectively resolves these issues and maintains high-quality results even when generating relatively small characters.
Regarding Chinese visual text generation, CharGen outperforms AnyText in addressing problems such as dropped strokes, added strokes, and inaccurate characters, as illustrated in Fig.~\ref{any-bench}. Notably, our method achieves superior results compared to previous methods when generating complex visual text, as shown in the last column of the comparison.

Additionally, we demonstrate the capability of our approach in editing tasks for both Chinese and English, achieving good context-fitting and accurate visual text generation. We also showcase the ability of visual text editing in the images presented in Fig.~\ref{edit-img}. Compared to AnyText, our method edits the visual text in images more accurately and integrates seamlessly with the background.

\subsection{Ablation Study}
In the ablation experiment, we build a small dataset of 200K images extracted from AnyWord-3M, comprising 160K Chinese images and 40K English images. Following the approach of AnyText, we also use the Wukong test set from the AnyText-benchmark to evaluate the effectiveness of each sub-module of CharGen. Given the small size of the training set, each model is trained for 20 epochs over 13 hours using 8 Tesla A100 GPUs. To assess the effectiveness of each module in our method, we conduct ablation studies on various settings as shown in Table~\ref{tab:ablation strudy}.
\begin{itemize}
    \item \textit{Character-Level Embedding: } The findings from \textit{Exp.1} and \textit{Exp.5} underscore the significant impact of character-level embedding on enhancing the accuracy of visual text rendering. In these experiments, only visual embedding is employed, without the inclusion of text embedding or any perceptual loss. The character-level encoding module we introduced demonstrates a performance improvement of $7.5\%$ compared to box-level embedding, highlighting its efficacy in refining text rendering quality.
\end{itemize}
\begin{itemize}
    \item \textit{MultiModel Encoder: } Comparing \textit{Exp.4} to \textit{Exp.5}, it is evident that visual embedding is significantly more beneficial than text embedding. This observation underscores the importance of visual features in enhancing text rendering accuracy. Furthermore, the comparison between \textit{Exp.5} and \textit{Exp.8} highlights the effectiveness of the MultiModel Embedding Fusion module. The results indicate that the character-level MultiModel Embedding Fusion module surpasses the performance of the visual encoder module alone by $3.7\%$. This improvement demonstrates the value of integrating both visual and text embeddings for comprehensive feature representation and enhanced model performance.
\end{itemize}
\begin{itemize}
    \item \textit{CharGen Perceptual Loss: } The results from \textit{Exp.2} to \textit{Exp.3} and \textit{Exp.6} to \textit{Exp.7} clearly demonstrate the superiority of CharGen Perceptual Loss over OCR Perceptual Loss, with accuracy improvements of $2.4\%$ and $3.4\%$, respectively. Moreover, \textit{Exp.5} and \textit{Exp.7} highlight that integrating CharGen Perceptual Loss enhances performance by $6.7\%$ compared to one without any Perceptual Loss supervision. Additionally, the findings from \textit{Exp.9} to \textit{Exp.11} suggest that the loss coefficient $\lambda$ achieves optimal performance at a value of 0.01.
\end{itemize}

\section{Conclusion and Limitation}

In this paper, we present CharGen, a highly accurate character-level visual text generation method. To tackle the challenges of generating complex visual text, particularly issues with inaccurate characters and strokes, we propose a character-level multimodal encoder. This encoder incorporates rich, fine-grained information, enhancing the model's ability to produce precise visual text. Furthermore, we introduce CharGen perceptual loss, a supervision loss with finer granularity, which significantly improves the accuracy of visual text details.
CharGen has been evaluated on two public benchmarks, where it surpasses existing methods and achieves state-of-the-art results on both Chinese and English test sets. Despite its high accuracy, our method has some limitations, such as generating redundant text in the background and reduced accuracy when dealing with extremely small-sized text. In future work, we aim to enhance the quality of rendering extremely small-sized text, suppress redundant background text, and enable text generation with controllable attributes.

{
    \small
    \bibliographystyle{ieeenat_fullname}
    \bibliography{main}
}


\end{document}